**RESEARCH ARTICLE**

# Probabilistic selection and design of concrete using machine learning

Jessica C. Forsdyke[1] , Bahdan Zviazhynski[2] , Janet M. Lees[1] and Gareth J. Conduit[2]

[1]Department of Engineering, University of Cambridge, Cambridge, United Kingdom
[2]Cavendish Laboratory, University of Cambridge, Cambridge, United Kingdom
**Corresponding author:** Jessica C. Forsdyke; Email: jf580@cam.ac.uk

**Abstract**

Development of robust concrete mixes with a lower environmental impact is challenging due to natural variability in constituent materials and a multitude of possible combinations of mix proportions. Making reliable property predictions with machine learning can facilitate performance-based specification of concrete, reducing material inefficiencies and improving the sustainability of concrete construction. In this work, we develop a machine learning algorithm that can utilize intermediate target variables and their associated noise to predict the final target variable. We apply the methodology to specify a concrete mix that has high resistance to carbonation, and another concrete mix that has low environmental impact. Both mixes also fulfill targets on the strength, density, and cost. The specified mixes are experimentally validated against their predictions. Our generic methodology enables the exploitation of noise in machine learning, which has a broad range of applications in structural engineering and beyond.

**Impact Statement**

This article demonstrates that machine learning can be used to predict the properties of concrete even with a sparse and noisy dataset. This has important applications to performance-based specification of concrete mixes—enabling appropriately durable and strong concretes to be specified while minimizing embodied carbon or cost. In cases where time-consuming and costly trials are required, this is particularly beneficial. The machine learning methodology developed and demonstrated in this article has application to the broader field of accelerated materials design, allowing bespoke materials to be designed rapidly for each particular application. Furthermore, there are many examples of other verticals where information is embedded in noise, including autonomous vehicles, additive manufacturing, and information engineering, where machine learning offers the opportunity to accelerate development, understanding, and impact.

## 1. Introduction

Concrete is the most heavily used construction material in the world. The only substance consumed in greater quantities is water (Sedgwick, 1991). Concrete is ideal for construction because it is readily

---

J.C.F. and B.Z. contributed equally to this work.

This research article was awarded an Open Data badge for transparent practices. See the Data Availability Statement for details.

adaptable—material properties including density, strength, durability, and appearance can be manipulated by adjusting the proportions and materials in the mix design.

Conventional concrete mixes comprise three primary constituents: cement, water, and aggregate. Cement acts as the binder in concrete and is both economically and environmentally costly (Wassermann et al., 2009). Production of cement is responsible for an estimated 5–6% of global $CO_2$ emissions (United Nations Framework Convention on Climate Change (UNFCCC), 2018), owing to both process emissions and energy demands (Allwood and Cullen, 2012). Therefore, the bulk of concrete is comprised of aggregates, which can be classified as either "fine" (sand), or "coarse" (gravel and rock). Once concrete is mixed and placed in a mold, compounds in the cement undergo hydration reactions with water (Neville, 2011), forming a complex microstructure of pores, hydrated cement paste, and aggregates. This gives the concrete its "hardened state" properties including resistance to both applied loads (strength, stiffness), and aggressive substances (durability).

In this era of increasing environmental $CO_2$ concentrations, the interaction of $CO_2$ with the built environment is of increasing interest (Talukdar and Banthia, 2016; Jiang et al., 2018). In particular, there exists a chemical reaction of carbon dioxide ($CO_2$) with hydration products in concrete (Papadakis et al., 1989), which is referred to as concrete carbonation. The $CO_2$ diffuses into the concrete from the surrounding environment. The resulting carbonated material has a changed microstructure compared to noncarbonated concrete (Groves et al., 1991; Greve-dierfeld et al., 2020), and a lower alkalinity of the pore water (Papadakis et al., 1989). Therefore, this carbonation can lead to structural damage since this loss of alkalinity facilitates corrosion of the internal steel reinforcement which is required to carry tensile loads in concrete structures (Page, 2007). Ultimately, carbonation is one of the primary deterioration mechanisms of steel-reinforced concrete structures, making it a significant area of concern.

Specification of the most appropriate concrete for a particular application can be carried out using a performance-based or prescriptive approach. Where a performance-based concrete specification process is implemented, targets are placed on concrete output properties, such as strength or carbonation coefficient. This is opposed to a prescriptive concrete specification approach, which limits the input parameters such as cement content or water/cement ratio (the principle applied by Eurocode BS EN 206:2013+A1 (British Standards Institution, 2016). At present, concrete mix design methods (Abrams, 1918; ACI Committee 211, 1991; Teychenné et al., 1997; Wilson and Kosmatka, 2011) use empirically derived relationships between mix ratios and concrete properties, in particular the strength of the concrete, to proportion mixes. In any large-scale concrete application following a performance-based approach, a variety of mixes are designed using one of these methods and then trial mixes are cast to verify that the final properties of a particular mix are acceptable. Performance-based specification can, therefore, lead to more efficient, optimal mixes being selected (Wally et al., 2022). However, the trial-and-error process can be both time-consuming and costly.

Machine learning offers an opportunity to capture complex multi-dimensional relationships between the inputs (mix proportions) and outputs (material properties), and to reduce the need for a trial-and-error approach in concrete design. Machine learning uses historical data to train a model, which later can be used to predict quantities of interest. This approach has been used to predict properties of many different materials (Bhadeshia et al., 1995; Sourmail et al., 2002; Agrawal et al., 2014; Ward et al., 2016; Ward et al., 2017; Kim et al., 2018), including concrete (Taffese et al., 2015; Ben Chaabene et al., 2020; Prayogo et al., 2020; Liu et al., 2021; Tran et al., 2022), proving its capability and versatility. Here we turn to machine learning to design concrete.

Machine learning methods can also utilize the uncertainty in predictions of target variables to focus on the most robust predictions. For example, the use of uncertainty has been extensively demonstrated and experimentally verified for design of materials most likely to fulfill target criteria (Conduit et al., 2017, 2018, 2019). Furthermore, values of uncertainty itself can be useful for predicting the quantity of interest (Goujon, 2009; Zerva et al., 2017; Zhang, 2020; Zviazhynski and Conduit, 2022). In concrete, the appearance of randomly distributed aggregates of different shapes and sizes may be considered similar to white noise, leading to variability and uncertainty in properties such as carbonation depth and compressive strength. Information can be extracted from uncertainty in some of the measured properties, and used to predict other properties to allow better insights and, ultimately, better mix design. In this work, we develop a methodology capable of extracting information from uncertainty. We use the methodology to select concrete mixes that have specific desired properties for sustainability and durability targets, providing a template for future machine learning applications to systems where there is information hidden in the noise.

In this article, we first explore the relationships between measured properties of concrete mixes prepared in a laboratory study in Section 2. We then describe the machine learning methodology capable of extracting information out of noise to help predict properties of concrete in Section 3. We next describe the application of this methodology to concrete specification in Section 4 and validate our proposal on experimental results in Section 5. Finally, we discuss the future applications of the methodology to concrete-related areas and beyond in Section 6.

## 2. Relationships Between Concrete Properties

To build our machine learning model on a solid foundation we first explore the relationships between various concrete properties using empirically collected data from 21 laboratory concretes, including two mixes reported in Forsdyke and Lees (2021b). The dataset (provided in the Supplementary Material) contains information about the proportions of constituent materials (cement, gravel, sand, and water) in each batch, as well as properties of interest (carbonation coefficient, environmental impact, strength, density, and cost) for the resulting concrete. In all concretes in this publication, the gravel used is crushed limestone with the maximum particle size of 10 mm, and the sharp sand used has a maximum particle size of 4 mm (with 80% of sand particles smaller than 1 mm).

The mixes were originally proportioned using the BRE method (Teychenné et al., 1997), which uses an estimated mass per cubic meter, also given. As a measure of concrete durability, 16 of the 21 concretes were exposed to a 4% $CO_2$ environment. The carbonation depth was measured over time from a minimum of three samples of each mix to produce a carbonation profile from which carbonation coefficient, $K$, was calculated. A low carbonation coefficient represents high resistance to carbonation. Another variable, preconditioning time, refers to the period between water curing and elevated $CO_2$ exposure for those specimens where carbonation performance was measured. An estimate of environmental impact in the form of kg embodied $CO_2$ per kg of concrete (kgeCO$_2$/kg) was also provided, as well as: compressive cube strength of the concrete; water saturated density measured in air; and estimated cost per kg (£/kg).

The correlations between the properties from the concrete dataset, measured with the Pearson correlation coefficient, are shown in Figure 1. The first nine rows represent input properties such as mix proportions and estimated mass per cubic meter assumed before the concrete has been cast. The final five rows represent measured properties of the resultant concrete. Correlations of particular interest are highlighted by the boxes (a–k) in Figure 1. Overall, Figure 1 shows that there are correlations between many of the variables in the concrete dataset, which can be exploited to fit a machine learning model to the data and make predictions. We now discuss notable correlations for each individual property.

### 2.1. Carbonation coefficient

A strong negative correlation between carbonation coefficient and cement content is observed at (a) in Figure 1. This is because higher proportions of cement provide higher volumes of carbonatable material in the concrete matrix once hydrated (Papadakis et al., 1991), increasing the time for carbonation to reach a particular depth. Strong positive correlations are observed between carbonation coefficient and crushed gravel content, (b), and between carbonation coefficient and total aggregate/cement ratio, (c). This is also logical when considering a fixed volume of hydrated cement paste will penetrate deeper into a concrete when more aggregates are suspended in it.

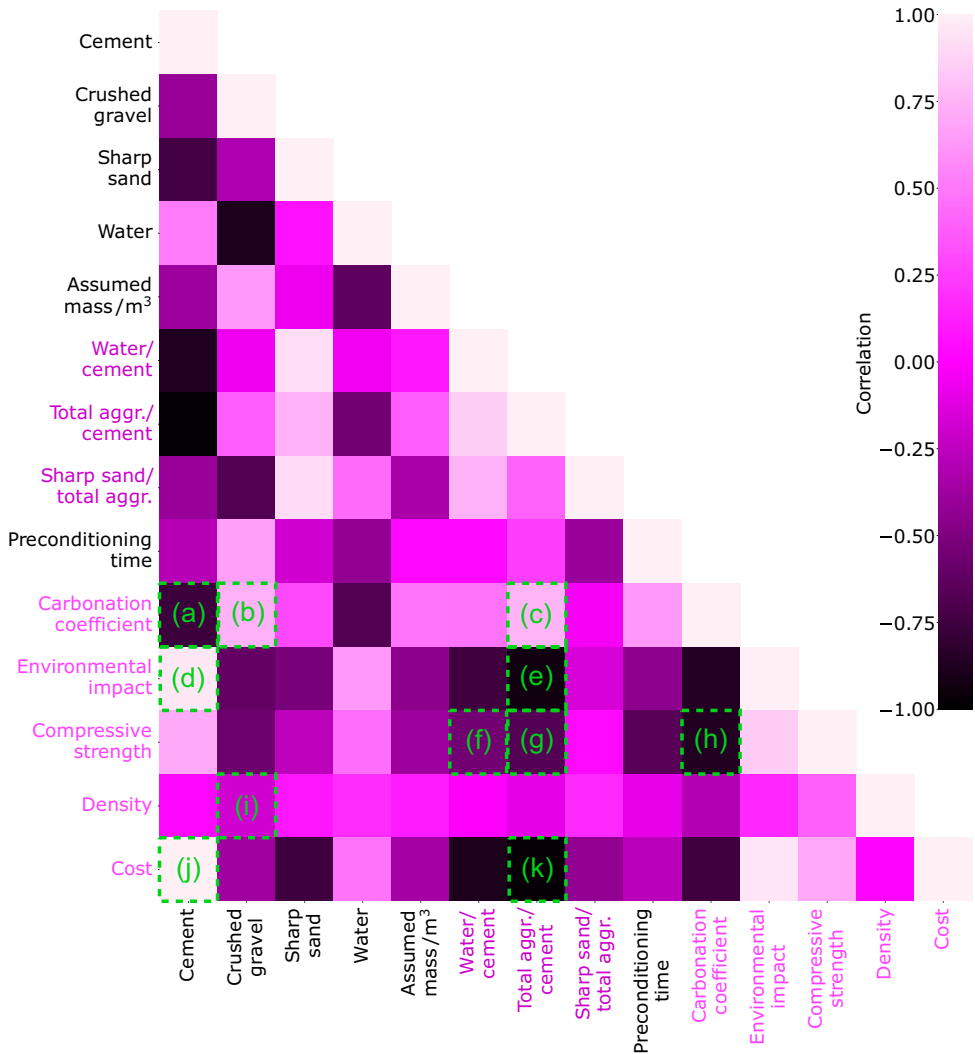

***Figure 1.*** *Correlation map for concrete properties in the training dataset. Light colors correspond to strong positive correlations, dark colors correspond to strong negative correlations, and intermediate colors correspond to weak correlations, as per the color scale shown. Green boxes highlight notable correlations. Properties written in dark pink are intermediate quantities derived from concrete mix proportions and properties written in light pink are target variables.*

### *2.2. Environmental impact*

It is well established that cement is the component of concrete responsible for the majority of its environmental impact. Whereas, the relative embodied carbon of the aggregates, acting as inert filler materials (Neville, 2011), are low. This is confirmed in Figure 1 by the strong positive correlation between environmental impact and cement content, (d), and the strong negative correlation between environmental impact and total aggregate/cement ratio, (e). With such strong correlations, we expect machine learning predictions for environmental impact to be more accurate than for other target variables, which have weaker correlations with composition variables.

 Environmental impact in this study is calculated assuming constant embodied emissions for each constituent component. Therefore, environmental impact is a linear function of concrete mix proportions. Generally, however, embodied emissions of each component depend on the manufacturing process, transport to the site, and solid waste generation (Babor et al., 2009; Sousa and Bogas, 2021). The effect of the manufacturing process, transportation, and solid waste generation on environmental impact cannot be evaluated analytically and so this study represents a template of how machine learning could be used to predict environmental impact.

### 2.3. Compressive strength

Figure 1 also demonstrates negative correlations between compressive strength and water/cement ratio, (f), and between compressive strength and total aggregate/cement ratio, (g). Concrete strength is widely considered to be a function of pore structure (Powers, 1958; Pantazopoulou and Mills, 1995; Chen et al., 2013). Above the minimum required for full cement hydration (Powers, 1958; Aïtcin, 2016), increasing the water/cement ratio leads to loss of concrete strength due to microscopic pores formed by water molecules that are not chemically bound in hydration products (Pann et al., 2004). Increasing the total aggregate/cement ratio similarly decreases strength due to loss of bond between the cement matrix and aggregates (Poon and Lam, 2008) and may change the degree of compaction (Marar and Eren, 2011) that leads to lower strength (Mindess et al., 2003). A strong negative correlation between strength and carbonation coefficient is also observed (h), supporting the validity of models that use strength to predict carbonation behavior (Silva et al., 2014; Forsdyke and Lees, 2021a).

### 2.4. Density

Density does not appear to have any strong correlation to the input parameters, which we envisage to make the application of machine learning to predict density more difficult than the other target variables. The strongest relationship observed is a negative correlation to the absolute volume of coarse aggregate, crushed gravel, (i). Density of concrete in the hardened state is influenced by the particle-size distribution having allowed effective compaction in the wet state (Sims et al., 2019). Total aggregate content varies minimally between the mixes in this study. Therefore, higher quantities of coarse aggregate correspond to lower quantities of fine aggregate within the data set (as seen by the strong negative correlation between sharp sand and crushed gravel). It is possible that this leads to less optimal particle packing, which in turn leads to the lower density observed here for mixes with higher coarse aggregate volumes. Other information not captured here, such as the particle size distribution and smoothness of the aggregates, may also influence density, leading to noise in this data.

### 2.5. Cost

A strong positive correlation is observed between cost and cement content, (j), whereas a strong negative correlation is observed between cost and total aggregate/cement ratio, (k). This is expected, since cement is the most expensive component of concrete, whereas aggregates act as a filler material and are much cheaper components of concrete. Like for environmental impact, we expect machine learning predictions for cost to be more accurate than for other target variables.

Cost in this study is calculated assuming fixed price for each constituent component. Therefore, cost, like environmental impact, is a linear function of concrete mix proportions. Generally, however, the price of each component depends on regulation, inflation, and supply (Velumani and Nampoothiri, 2018; Ma et al., 2022). The effect of these factors on cost cannot be evaluated analytically and so this study serves as a foundation showing how machine learning could be used to predict cost.

## 3. Machine Learning Methodology

Machine learning algorithms are trained on preexisting data to make predictions. In this case, the predicted quantities are the properties of concrete mixes. A few examples of widely used machine learning algorithms are *k*-means clustering (Hastie et al., 2001), neural networks (Heskes, 1997), and Gaussian processes (Tancret, 2013). In this article, we use the random forest algorithm (Pedregosa et al., 2011), as it is computationally cheap and robust for finding nonlinear relationships. We compare a single-layer linear model with a single-layer random forest model and our two-layer random forest model that can extract information from the noise present in concrete data. Figure 2 shows the flow of information through these three model types.

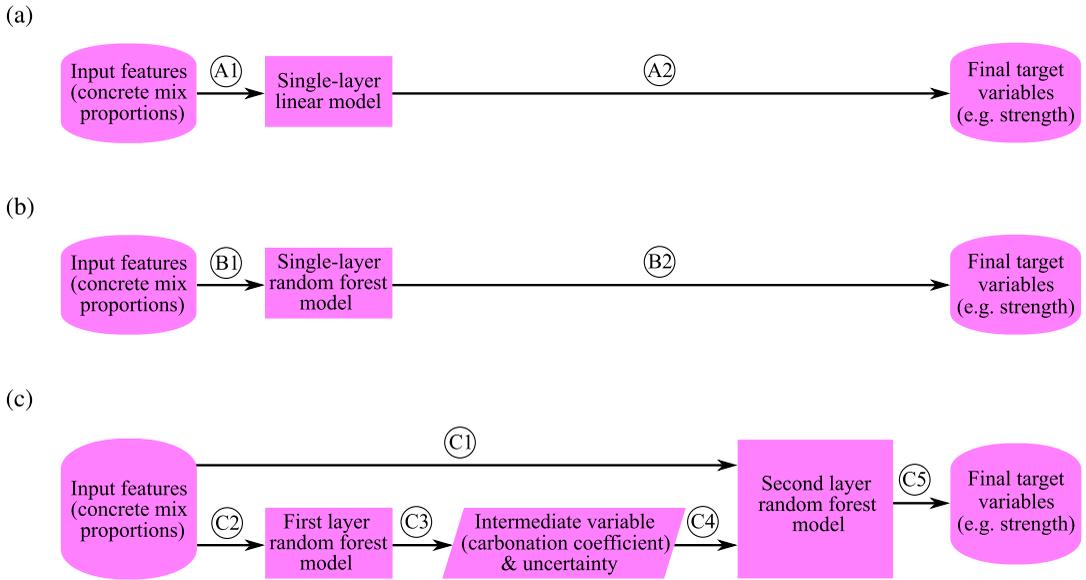

**Figure 2.** *(a) Flowchart for the single-layer linear model. (b) Flowchart for the single-layer random forest model. (c) Flowchart for the two-layer random forest model.*

### *3.1. Single-layer linear model*

The most basic method for property prediction is to fit a hyperplane to the training data. The hyperplane uses the input values (e.g., concrete mix proportions) to directly predict the final target variable (e.g., strength). Figure 1 showed the strength of linear correlations between input and output properties, motivating this approach. This method, which we call a single-layer linear model, is illustrated by Flows A1 and A2 in Figure 2a.

The method is simple and robust; however, it does not capture the nonlinear relationships between concrete properties. If the final target variable has a nonlinear dependence on the inputs, the linear model predictions will not be accurate. Therefore, we also consider a model that captures nonlinearities in the data, which is a single-layer random forest model.

### *3.2. Single-layer random forest model*

The random forest model is a collection of independent identical regression trees (Loh, 2011). During the training phase, each regression tree learns how to map the input variables to the target variables. The random forest algorithm computes the prediction and its uncertainty by bootstrapping (Hastie et al., 2001). When bootstrapping a dataset with *N* entries, new subsets are generated by sampling *N* entries randomly, with replacement. Each subset is then used to train one regression tree in the random forest. The compound predictions are averaged to give the random forest prediction, and their standard deviation is the uncertainty in prediction.

The depth of regression trees in a random forest, and, therefore, accuracy of the predictions, is determined by the hyperparameters of the random forest. To achieve the best possible predictions for blind data we tune these hyperparameters, assessing the accuracy of predictions using leave-one-out cross-validation (Hastie et al., 2001) on the training data. For predictions of each property, we use the same hyperparameters to mitigate overfitting. The random forest algorithm can learn and exploit correlations between various target variables and their uncertainties to make more accurate predictions of the final target variable of interest.

A single-layer random forest approach follows Flows B1 and B2 in Figure 2b to use only the input features (concrete mix proportions) to predict the final target variables (e.g. strength).

### 3.3. Two-layer random forest model

Alternative to the conventional single-layer approach, the flow of information through the two-layer random forest model is demonstrated in Figure 2c. In the two-layer method, we train the first random forest model on the concrete mix proportions (Flow C1) to predict all of the intermediate and output variables such as carbonation coefficient, alongside its uncertainty (Flow C2), following the approach described in Zviazhynski and Conduit (2022). We then train the second layer random forest model, taking the concrete mix proportions (Flow C3) and other recently predicted variables including values for uncertainty (Flow C4) to predict the outputs, for example, strength (Flow C5).

Using an intermediate variable and its uncertainty as an input for the second layer random forest model is particularly useful if the intermediate variable is cheaper to measure than the target variable. For example, strength is both more economical to measure and more commonly measured in concrete than carbonation coefficient. Therefore, strength can be used as an intermediate variable in the two-layer random forest model to predict carbonation coefficient, without the need for time-consuming carbonation experiments. On the other hand, to assess the condition of an existing structure, taking a large core from the structure to measure compressive strength can be particularly destructive. Instead, an intermediate quantity, such as carbonation depth, can be measured from smaller, less invasive samples, to estimate the strength with less damage to the structure. The strategy of using the expected value of an intermediate variable to help predict the final variable has previously been successfully applied to materials and drugs design to exploit relationships between various properties (Verpoort et al., 2019; Irwin et al., 2020a; Mahmoud et al., 2021), but we now use uncertainty in the intermediate variable.

The uncertainty in the intermediate variable that machine learning should capture arises from the inevitable scatter in training data due to experimental uncertainty. There are two methods to quantify the uncertainty in the intermediate variable. The first method is to calculate the standard deviation of predictions of regression trees in the random forest, and this quantity is used in the two-layer random forest model. The second method is to calculate the standard deviation of the experimental measurements. In both cases, the uncertainty in a given variable has the same unit as this variable.

To demonstrate the utilization of uncertainty in the two-layer random forest model, we consider an example shown in Figure 3. Here, for a given value of cement content $X$, the first layer random forest model predicts the corresponding value of carbonation coefficient $Y$ and its uncertainty $\sigma_Y$, as shown in the leftmost plot in Figure 3. Then the second layer random forest model utilizes uncertainty in carbonation coefficient to predict the corresponding value of compressive strength, $Z$, as shown in the middle plot in Figure 3. Compressive strength is a decreasing function of uncertainty in carbonation coefficient (Spearman correlation of $-0.79$), which enables the two models to work together to predict compressive strength for a given cement content, as shown in the rightmost plot in Figure 3.

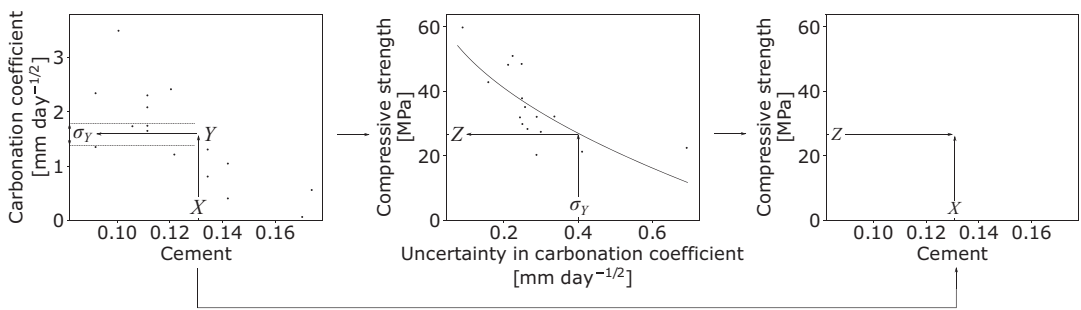

***Figure 3.*** *Utilization of uncertainty in the two-layer random forest model. Starting with cement content X, the model utilizes uncertainty in carbonation coefficient to predict compressive strength Z.*

### 3.4. Training and testing the model

First, the hyperparameters of the three alternative models were tuned using leave-one-out cross-validation. In this method, for each entry in the existing data on concrete mixes, the prediction of the concrete property (e.g., strength) is made by the model trained on the rest of the existing data (Figure 4a). Then these predictions are compared against the true values to obtain the leave-one-out cross-validation $R^2$, coefficient of determination (ranges from 1 for perfect predictions to $-\infty$ for arbitrarily inaccurate predictions), which is to be maximized. The leave-one-out cross-validation $R^2$ values for each property after hyperparameter tuning of each model can be seen in the table in Figure 4b.

The single-layer linear model gives good predictions for environmental impact, compressive strength, and cost, since these are approximately linear functions. However, linear model predictions of carbonation coefficient and density are inaccurate. The single-layer random forest model improves on those predictions at the expense of environmental impact, strength, and cost prediction accuracy. As expected from the data analysis in Section 2, the highest values of cross-validation $R^2$ are achieved for environmental impact and cost by all three models and the lowest values are generally achieved for density. The two-layer model, capable of exploiting correlations between target variables and uncertainties in them, therefore outperforms the single-layer linear model and single-layer random forest model. This demonstrates the necessity of data-driven machine learning for property predictions and viability of the approach even when the dataset only contains 21 samples. The hyperparameters of single-layer linear model, single-layer random forest model, and two-layer random forest model can be found in the Supplementary Material.

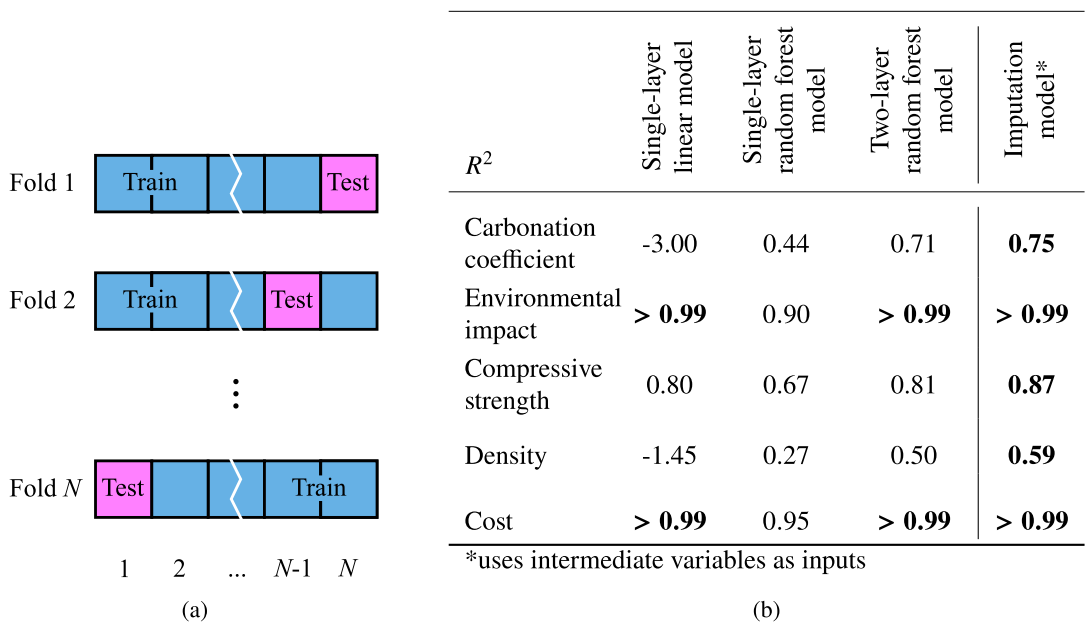

***Figure 4.*** *(a) Schematic of leave-one-out cross-validation. Blue squares are entries in the existing data and magenta squares are the test entries for each fold. (b) Table of leave-one-out cross-validation $R^2$ values for property predictions. Numbers in bold are the best $R^2$ values for a given property.*

To evaluate the effect of intermediate variables, we furthermore test a single-layer random forest model that has access to values of the intermediate variables taken from the experimental data to predict the target variable. This model is known as imputation model and has been successfully used in materials and drug discovery (Verpoort et al., 2019; Whitehead et al., 2019; Irwin et al., 2020a, 2020b; Mahmoud et al., 2021). Imputation models exploit correlations between the intermediate variables and the target variable and therefore achieve higher cross-validation $R^2$ than both a single-layer linear model and a single-layer random forest model. The imputation model does not need to predict intermediate variables so does not propagate the error in them, which leads to better cross-validation $R^2$ than the two-layer random forest model. However, the imputation model requires measured concrete properties as inputs, so is unsuitable for designing and predicting properties of a novel concrete mix. Therefore, for the remainder of this study, we adopt the two-layer random forest model.

The two-layer random forest model has access to cement type, concrete mix proportions, ratios of concrete mix proportions, assumed mass/m$^3$, and preconditioning time. The relative importances of each feature for predicting the corresponding target variable are presented in the Supplementary Material. After having trained the two-layer random forest model with the tuned hyperparameters on all of the existing data on concrete mixes, the model can now be used to predict the target variables and uncertainties in them for the unseen mixes. The probabilities of the unseen mixes satisfying the set target criteria would then be calculated; the mixes with the highest probability of satisfying the given targets would be experimentally validated.

## 4. Concrete Specification

With the machine learning model in place, we are well-positioned to explore completely unseen concrete mixes. We below specify two sets of challenging target properties required of the concrete. To emulate real-life usage on a construction site, where local material availability will determine the composition of possible mix designs, we use machine learning to seek the best mix from a family of hypothetical mixes, over a grid of water/cement ratio values. We first specify a total water content of 205 kg/m$^3$, chosen assuming constant workability requirements of these mixes. The hypothetical mixes are then proportioned following the BRE method of mix design (Teychenné et al., 1997), which uses the relative density of the aggregates (2.7 in this case) and the percentage of fine aggregate passing through a 600 micron sieve (70% in this case) to calculate the proportions of the remaining constituents and total density of the concrete for each water/cement ratio. This ensures the proportions of all potential mixes are realistic, and the mixes are viable. Machine learning is then used to explore this space of mixes, and allows for the prediction of extensive properties of these mixes which would otherwise need to be experimentally obtained.

For the first target mix, Low-$K$, we focus on minimizing carbonation coefficient and therefore providing high resistance to carbonation. This mix would be ideal for structural applications with severe exposure conditions, where a high resistance to carbonation is required to protect steel reinforcement from carbonation-induced corrosion.

For the second target mix, Low-$E$, we focus on minimizing environmental impact whilst maintaining a reasonably low carbonation coefficient, which is challenging as these properties are negatively correlated. This mix would be ideal for large-scale structural applications with moderate exposure conditions, where low environmental impact is desired.

For both scenarios, we also seek to satisfy constraints on minimum strength, maximum density, and maximum financial cost. A summary of the two sets of target criteria is presented in Table 1. We now calculate the probability of successfully meeting the target criteria for each hypothetical mix so that we can focus on the most robust mixes that will work in practice. First, for each target variable (carbonation coefficient, environmental impact, compressive strength, density, and cost) of a given hypothetical mix, we construct the probability distribution curve of machine learning predictions that reflects the uncertainty in the prediction. Then, we evaluate the area under this curve within the target region. Finally, we

**Table 1.** *Target criteria applied when selecting from available mix designs.*

| Target criteria | Low-K | Low-E |
| --- | --- | --- |
| Carbonation coefficient (mm day$^{-1/2}$) | < 1.2 | < 2.4 |
| Environmental impact (kg $CO_2$e/kg) | < 0.150 | < 0.108 |
| Compressive strength (MPa) | > 30 | > 20 |
| Density (kg/m$^3$) | < 2,350 | < 2,350 |
| Cost (£/kg) | < 0.028 | < 0.028 |

*Note.* Chosen mixes are those most likely to satisfy all of the criteria.

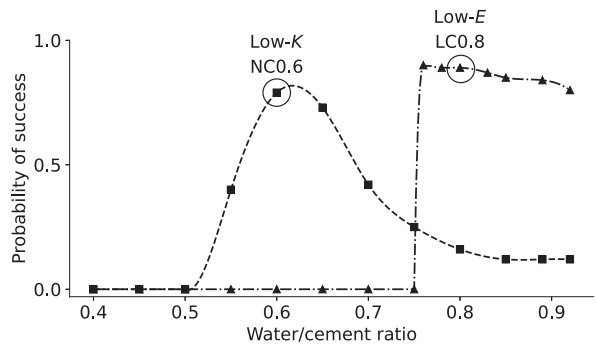

**Figure 5.** *Probability of success of the family of hypothetical mixes satisfying the Low-K (dashed line, squares) and Low-E (dash-dotted line, triangles) criteria, plotted against water/cement ratio. The selected mixes are circled.*

multiply the areas obtained for each target variable to give the probability of the given mix successfully meeting the target criteria. We seek the mix that maximizes the probability of success.

Figure 5 demonstrates the probability of successfully meeting the target criteria for the family of hypothetical mixes with water/cement ratios ranging from 0.4 to 0.95. The water/cement ratio in particular is varied as this was shown in Figure 1 to be an important variable in the mix parameters. The lines are plotted using cubic spline fit.

From the potential mixes, the 0.6 water/cement ratio mix (NC0.6) and the 0.8 water/cement ratio mix (LC0.8) were found. In general, increasing water/cement ratio improves the environmental impact, whereas reducing the ratio improves the carbonation resistance. A compromise between these two factors that fulfills the Low-K criteria outlined in Table 1 with high probability of success is achieved by NC0.6. For the Low-E criteria, the accurate model for environmental impact (see the table in Figure 4b) with small uncertainty drives the sudden increase in probability of success seen in Figure 5 above a water/cement ratio of 0.75. LC0.8 is selected to be sufficiently far from the boundary to circumvent experimental error in mixing whilst retaining a high probability of success. The two selected mix designs are given in Table 2.

## 5. Experimental Validation

### 5.1. Manufacturing

To validate the properties of the two proposed mixes, several 100 mm × 100 mm × 100 mm cubes of these mixes were cast. The mixes used crushed limestone gravel with a maximum particle size of 10 mm, and sharp sand with a maximum particle size of 4 mm and 80% of particles smaller than 1 mm. These were the same constituent materials used for the concretes in the training set. The concrete was mixed for a total of 5 min in a pan mixer, before being transferred to oiled plastic cube molds. Whilst filling, the molds were vibrated on a vibrating table for a total of 12 s to ensure full compaction of the concrete, and then skimmed with a trowel to achieve a flat exposed face. The concrete cubes were removed from the molds following an initial setting period of 24 hr, during which they were covered with polythene sheeting to prevent moisture loss, and then cured under water at 20°C until 28 days old. During casting, a trace amount (0.2% by mass) of dye was added to the LC0.8 mix for experimental reasons. We measured carbonation coefficient, compressive cube strength, and water-saturated density in air. Cost and embodied carbon of each of these mixes was calculated using coefficients for each of the constituent materials from a published cost estimation spreadsheet (Fibo Intercon, 2019) and the ICE database (Jones and Hammond, 2019) respectively. Experimental uncertainty of 0.7% is assumed for measurements of mass (deriving from a 0.1 kg precision of weighing scales for a mass measurement of 15 kg). For other experimental values, uncertainty is quantified through repeat measurements.

**Table 2.** *The two compositions that are each most probable to fulfill their respective target criteria, so are proposed for experimental validation.*

| Target criteria | Low-*K* | Low-*E* |
| --- | --- | --- |
| Selected mix | NC0.6 | LC0.8 |
| Probability of success | 0.79 | 0.89 |
| Cement (CEM I 52.5N) (%) | 14.2 | 10.5 |
| Crushed gravel (%) | 48.9 | 48.2 |
| Sharp sand (%) | 28.4 | 32.6 |
| Water (%) | 8.5 | 8.5 |
| Water/cement ratio | 0.6 | 0.8 |
| Total aggregate/cement ratio | 5.5 | 7.7 |

### 5.2. Validation of individual properties

In this section, we measure and validate each of the five properties of the two concrete mixtures. The machine learning predictions of the five properties (carbonation coefficient, environmental impact, strength, density, cost) and their experimental values, for both concrete mixes, are presented in Figure 6. For both concrete mixes, all the predictions agree with the corresponding experimental values within standard error. We now discuss the measurement and validation of each individual property.

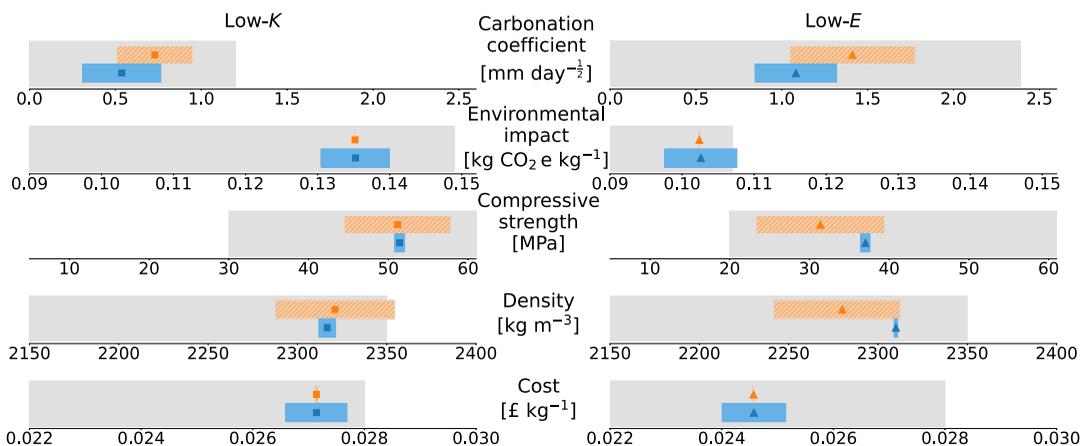

**Figure 6.** *Summary of machine learning predictions (orange, hatched) and experimental results (blue) of properties for the two concrete mixes. Bars correspond to standard error regions for both predicted and experimental values. Gray areas correspond to the property targets.*

### 5.2.1. Carbonation coefficient

Carbonation reactions take place in the pores of the cement paste matrix when concrete is exposed to $CO_2$. The process is modeled as 1-dimensional diffusion according to Fick's first law (Kropp and Hilsdorf, 1995), where the square of the depth of penetration of $CO_2$, known as the carbonation depth, $x$, is proportional to the exposure time, $t$, by a carbonation coefficient, $K$, which is itself a function of the concrete's properties and the concentration of environmental $CO_2$. Including all boundary conditions, the relationship is defined (Moreno, 2013):

$$x(t) = \sqrt{x(0)^2 + K^2 t}, \tag{1}$$

where $t$ is the time of exposure to a constant concentration $CO_2$ and $x(0)$ is the initial carbonation depth at $t = 0$ (often equal to 0 mm).

Since the carbonation process in extremely slow at atmospheric concentrations, taking years to reach significant carbonation depths, accelerated carbonation tests are performed under elevated $CO_2$ concentrations to gauge the relative performance of different concrete mixes. A test concentration of 4% is typical, resulting in the value of $K$ herein referred to as the 4% accelerated carbonation coefficient, $K_{4\%}$.

Noncarbonated material is revealed on the freshly split concrete surface in Figure 7 in pink, due to a phenolphthalein indicator solution. Carbonated material remains gray. Aggregates are generally of low permeability to $CO_2$ compared to the matrix of cement paste and pore space. Therefore, they impede the progression of carbonation through the cement paste and result in a nonuniform tortuous carbonation front (Huang et al., 2012; Shen and Pan, 2017) seen in Figure 7. To account for this tortuosity, the carbonation depth is measured using the method outlined in BS 1881-210:2013 (BSI, 2013) at multiple equidistant locations along the front. The carbonation depth at exposure time $t$, $x(t)$, is taken as the average of these measurements, shown in Figure 7.

The carbonation depth data was converted into a 4% accelerated carbonation coefficient, $K_{4\%}$ by curve fitting all available data points of $x(t)$ versus $\sqrt{t}$ for each mix to equation (1), shown in Figure 8, using the curve fitting method from BS EN 12390-12:2020 (BSI, 2020). The error in carbonation coefficient arises from variability of $x(t)$ across the carbonation front. To estimate this error, we curve fit $x(t) + \sigma_x(t)$ versus $\sqrt{t}$, where $\sigma_x(t)$ is standard deviation of carbonation depth at a given $t$, and obtain the upper bound for carbonation coefficient. We then subtract the $K_{4\%}$ value from the upper bound to estimate the error in carbonation coefficient. The same estimate can be obtained using the lower bound for carbonation coefficient, as the latter is assumed to be normally distributed.

It can be seen in Figure 6 that the machine learning predictions of carbonation coefficient agree with the experimentally measured values within standard error. We note that machine learning predictions are higher than the corresponding experimental values for both mixes, this may be due to prevalence of high carbonation coefficient values in the training data.

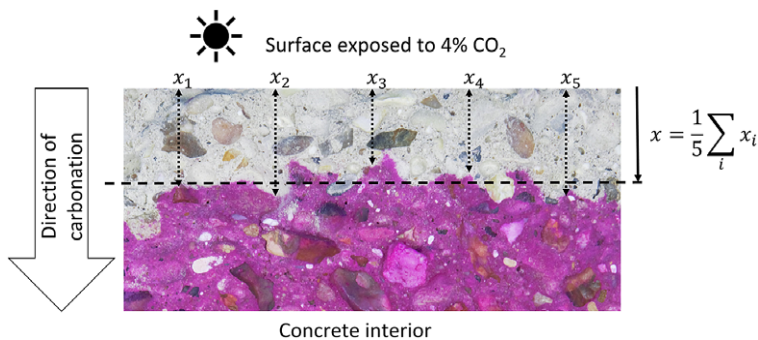

**Figure 7.** *Concrete sample with aggregate/cement ratio of 6.9, with upper surface exposed to 4% $CO_2$ for 49 days, other surfaces contact the rest of the sample. The carbonation front is revealed using 1% phenolphthalein in ethanol indicator solution (magenta when not carbonated).*

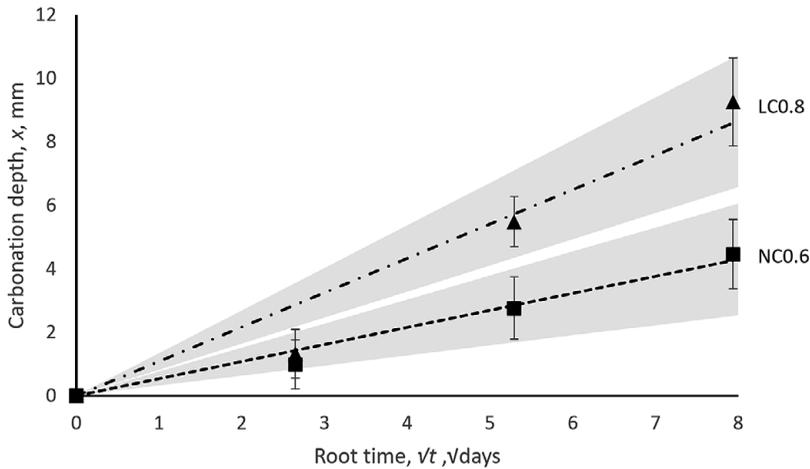

*Figure 8.* Experimental carbonation results showing estimate of carbonation coefficient (black linear fit) and standard error bounds (gray regions).

*5.2.2. Environmental impact*
The environmental impact of the newly proposed concrete mixes was estimated with equation (2):

$$\mathrm{kg\,CO_2e/kg_{concrete}} = \sum_i e_i f_i, \qquad (2)$$

where $e_i$ is the embodied $CO_2$ per kg of the $i^{\mathrm{th}}$ material, and $f_i$ is the mass fraction of the $i^{\mathrm{th}}$ material in the concrete mix.

The coefficients for embodied $CO_2$ in each of the constituent materials, $c_i$, were taken from the ICE database (Jones and Hammond, 2019) and presented in the Supplementary Material. Experimental error in this value is estimated by assuming an error in mass measurements of 0.7%, based on precision of scales used for the experimental series of $\pm$ 0.1 kg for every 15 kg weighed.

Figure 6 shows that the machine learning predictions of environmental impact are in excellent agreement with the experimental values. This is due to the fact that environmental impact is an approximately linear function of the concrete composition proportions and therefore straightforward to predict.

*5.2.3. Strength*
For each of the newly proposed concrete mixes, following a wet curing period of 28 days, compressive cube strength was measured in accordance with BS EN 12390-3:2019 (BSI, 2019a) on three 100 mm × 100 mm × 100 mm cube specimens, giving a mean achieved cube strength and standard deviation when assuming that cube strengths are normally distributed (Dayaratnam and Ranganathan, 1976). This is the property most commonly measured from field concretes, as it determines the load-bearing capacity of structures made from this mix.

It can be seen in Figure 6 that the machine learning predictions of strength agree with experimentally measured values within standard error. Machine learning predictions are lower than the corresponding experimental values for both mixes due to prevalence of lower strength values for the mixes with similar cement content in the training set.

*5.2.4. Density*

The water-saturated density in air of concrete was measured using the method outlined in BS EN 12390-7:2019 (BSI, 2019b) for cubes of 100 mm × 100 mm × 100 mm. Three cubes of each mix were measured, and the standard deviation of these measurements recorded as the experimental error. Low experimental error indicates that the concrete is well mixed in the fresh state, which should also reduce variability of other properties.

It can be seen in Figure 6 that the machine learning predictions of density agree with experimentally measured values within standard error. This is a particular success because the leave-one-out $R^2$ value for the two-layer model in Figure 4b was lower than for the other properties. The machine learning prediction is higher than the experimental value for the NC0.6 Low-$K$ mix and lower than the experimental value for the LC0.8 Low-$E$ mix, that is, machine learning predictions have larger variance than the experimental values. This may be because density varies with the aggregate size, which can be different even for similar mixes, making it difficult to predict density.

*5.2.5. Cost*

The cost of concrete is calculated using equation (3). Experimental error in this value is estimated using the same assumptions as experimental error in environmental impact.

$$£/\text{kg}_{\text{concrete}} = \sum_i c_i f_i, \tag{3}$$

where $c_i$ is the price per kg of the $i^{\text{th}}$ material, and $f_i$ is the mass fraction of the $i^{\text{th}}$ material in the concrete mix.

Commercially, the pricing of concrete mixes will vary largely dependent on not only mix proportions but also scale of project or application. For this reason, representative values have been assumed from Fibo Intercon (2019) for medium size batches to give realistic relative values between concretes for optimization, but these should not be considered absolute. The values used are presented in the Supplementary Material. Experimental error in the cost is estimated by assuming an error in mass measurements of 0.7%, based on precision of scales used for the experimental series of ± 0.1 kg for every 15 kg weighed.

It can be seen in Figure 6 that the machine learning predictions of cost show excellent agreement with the experimental values. This is due to the fact that cost, like environmental impact, is an approximately linear function of the concrete composition proportions and therefore relatively easy to predict.

## 6. Discussion

### 6.1. Ramifications for concrete design and specification

The concretes proposed and experimentally validated in this work demonstrate the potential of the machine learning methodology to predict concrete behavior. Carbonation coefficient, environmental impact, strength, density, and cost are chosen as examples of constraints in concrete specification, but this could be extended to other properties, such as: fresh state behavior (e.g. slump, wet density), structural response (e.g. stiffness, flexural tensile strength), or durability behavior (e.g. permeability, porosity).

Performance-based specification is a growing area of research within concrete durability design, including design for resistance to carbonation (Younsi et al., 2011; von Greve-Dierfeld and Gehlen, 2016a, 2016b; Wally et al., 2022). Performance-based approaches may allow more economical and sustainable structures to be realized (Teplý and Vořechovská, 2009). However, to satisfy performance requirements, concrete properties must be demonstrated through tests such as the compressive strength and accelerated carbonation tests employed in this work. Using knowledge of the performance of previous specimens, machine learning could reduce the need for trial and error through such testing when selecting a mix, saving vital time on construction projects. This also enables the selection over multiple different parameters, including environmental impact, meaning that sustainability can be prioritized whilst still fulfilling other requirements. This could herald an era of just-in-time concrete design, with bespoke mixes specified on the construction site that offer optimal properties.

### 6.2. Future uses of machine learning

The machine learning method employed in this work can operate on any number of input features and target variables. The generic algorithm can be applied to images of concrete (Lemaire et al., 2005), where noise in the carbonation fronts or the distribution of aggregates may contain information about concrete properties and help improve their predictions. The method could also be used to predict the temporal dynamics of concrete properties. This would be useful to specify concrete mixes that satisfy the given targets throughout their life cycle.

The approach to extract information from noise has potential applications in areas beyond concrete design as well. One of these areas is autonomous vehicles. Here, uncertainty in the distance measured to the object contains information about its shape or type. For example, high uncertainty in distance tells that the object could be a fence, whereas low uncertainty could be characteristic of a wall. The methodology could also be applied to additive manufacturing (Rasiya et al., 2021), where noise in metal powder microstructure can be used to devise the optimal melting process. Another potential application area is information engineering, where noise in the data transmitted by sensors can be used for object tracking (Płaczek and Bernaś, 2014). Beyond engineering, the methodology could be applied to research of cancer, which is known to cause genetic chaos (Calin et al., 2003). The information extracted from this chaos could potentially be used for early cancer detection.

### 6.3. Conclusions

This work presented the use of a two-layer random forest regression model to select concrete mix designs with the highest probability of successfully achieving various target properties. The model extracts information out of noise, making it particularly applicable to the random distribution of aggregates within the concrete matrix that drives variability of concrete properties, such as strength and carbonation coefficient. The methodology was effective even when trained on sparse data, and gave leave-one-out cross-validation $R^2$ values above 0.50 for difficult-to-predict density and above 0.99 for environmental impact and cost, demonstrating overall superior performance than both a single-layer linear model and single-layer random forest model. Predictions for two blind mixes were experimentally validated to within standard error. Overall, these results are promising for future use of machine learning that can exploit noise for performance-based design of concrete across multiple properties, as well as for other materials and applications.

**Acknowledgments.** The authors acknowledge the assistance of lab technicians in the Civil Engineering building, University of Cambridge in conducting the experimental work reported here. There is Open Access to this article at https://www.openaccess.cam.ac.uk. For the purpose of open access, the authors have applied a Creative Commons Attribution (CC BY) license to any Author Accepted Manuscript version arising.

**Author contribution.** Conceptualization: J.C.F., B.Z., J.M.L., G.J.C.; Data curation: J.C.F.; Formal analysis: J.C.F., B.Z.; Funding acquisition: J.M.L., G.J.C.; Investigation: J.C.F.; Methodology: J.C.F., B.Z.; Project administration: J.M.L., G.J.C.; Software: B.Z.; Supervision: J.M.L., G.J.C.; Validation: J.C.F., B.Z.; Visualization: J.C.F., B.Z.; Writing—original draft: J.C.F., B.Z.; Writing—review and editing: J.M.L., G.J.C.

**Competing interest.** G.J.C. declares a potential financial conflict of interest as a Director of machine learning company, Intellegens Ltd. The other authors declare no competing interests exist.

**Data availability statement.** The data relating to this work are given in the Supplementary Material and can also be found at https://doi.org/10.17863/CAM.92245.

**Funding statement.** The authors acknowledge the financial support of the Engineering and Physical Sciences Research Council (Grant Nos. EP/N017668/1 and EP/N509620/1), Harding Distinguished Postgraduate Scholars Programme Leverage Scheme, and the Royal Society (Grant No. URF\ R\ 201,002). The funder had no role in study design, data collection and analysis, decision to publish, or preparation of the manuscript.

**Supplementary Materials.** The supplementary material for this article can be found at http://doi.org/10.1017/dce.2023.5

A.

*Table 1.* Training data. Values given are averages

| Mix | Cement type | Cement | Crushed gravel | Sharp sand | Water | Estimated mass/m$^3$ | Pre-conditioning time | Carbonation coefficient, $K_{4\%}$ | Environmental impact | Compressive strength, $f_{cube}$ | Water sat. density | Cost |
|---|---|---|---|---|---|---|---|---|---|---|---|---|
| | [CEM - ] | [%] | [%] | [%] | [%] | [kg/m$^3$] | [days] | [mm/day$^{1/2}$] | [kgCO$_2$e/kg] | [MPa] | [kg/m$^3$] | [£/kg] |
| C25 I | IIA 32.5 R | 11.1 | 52.8 | 28.1 | 7.8 | 2414 | 141 | 1.742 | 0.095 | 27.46 | 2337 | 0.025 |
| C25 II | IIA 32.5 R | 11.1 | 52.8 | 28.1 | 7.8 | 2414 | 138 | 2.301 | 0.095 | 21.30 | 2317 | 0.025 |
| C25 III | IIA 32.5 R | 11.1 | 52.8 | 28.1 | 7.8 | 2414 | 16 | 1.648 | 0.095 | 28.35 | 2300 | 0.025 |
| C25 IV | IIA 32.5 R | 11.1 | 52.8 | 28.1 | 7.8 | 2414 | 16 | 2.081 | 0.095 | 29.87 | 2297 | 0.025 |
| C20 I | IIA 32.5 R | 10.0 | 56.0 | 27.4 | 6.5 | 2448 | 141 | 3.493 | 0.087 | 22.49 | 2263 | 0.024 |
| C18 I | IIA 32.5 R | 12.0 | 52.9 | 27.3 | 7.8 | 2409 | 138 | 2.414 | 0.102 | 20.31 | 2260 | 0.026 |
| C30 I | IIA 32.5 R | 14.2 | 52.5 | 25.5 | 7.8 | 2409 | 16 | 1.045 | 0.119 | 35.14 | 2247 | 0.027 |
| C40 I | IIA 32.5 R | 17.4 | 51.4 | 23.4 | 7.8 | 2411 | 16 | 0.558 | 0.145 | 42.85 | 2220 | 0.029 |
| C45 I | I 52.5 N | 14.6 | 48.5 | 28.2 | 8.7 | 2344 | - | - | 0.139 | 49.90 | 2287 | 0.027 |
| C55 I | I 52.5 N | 13.7 | 51.5 | 27.2 | 7.6 | 2374 | - | - | 0.131 | 61.62 | 2333 | 0.027 |
| C50 I | I 52.5 N | 16.6 | 44.6 | 28.9 | 9.9 | 2312 | - | - | 0.157 | 53.48 | 2300 | 0.029 |
| C20 II | I 52.5 N | 10.3 | 47.7 | 33.2 | 8.7 | 2343 | - | - | 0.100 | 25.30 | 2237 | 0.024 |
| C35 I | I 52.5 N | 8.9 | 50.4 | 33.1 | 7.6 | 2374 | - | - | 0.088 | 38.76 | 2290 | 0.024 |
| C30 II | I 52.5 N | 9.2 | 50.9 | 32.4 | 7.3 | 2452 | 8 | 2.341 | 0.090 | 32.12 | 2240 | 0.024 |
| C45 II | I 52.5 N | 13.4 | 49.5 | 28.9 | 8.1 | 2421 | 8 | 1.304 | 0.128 | 48.26 | 2357 | 0.027 |
| C30 III | I 52.5 N | 9.2 | 50.9 | 32.4 | 7.3 | 2452 | 14 | 1.349 | 0.090 | 32.21 | 2343 | 0.024 |
| C45 III | I 52.5 N | 13.4 | 49.5 | 28.9 | 8.1 | 2421 | 14 | 0.806 | 0.128 | 48.53 | 2360 | 0.027 |
| C25 V | I 52.5 N | 10.6 | 48.3 | 32.1 | 9.1 | 2411 | 17 | 1.732 | 0.103 | 31.93 | 2310 | 0.025 |
| C30 IV | I 52.5 N | 12.2 | 48.8 | 29.9 | 9.1 | 2408 | 17 | 1.212 | 0.117 | 37.85 | 2307 | 0.026 |
| C45 IV | I 52.5 N | 14.2 | 48.9 | 27.9 | 9.0 | 2409 | 17 | 0.403 | 0.135 | 51.01 | 2330 | 0.027 |
| C55 II | I 52.5 N | 17.0 | 48.4 | 25.7 | 9.0 | 2409 | 17 | 0.062 | 0.161 | 59.83 | 2327 | 0.029 |

**B.**

*Table 2.* Hyperparameters of single-layer linear model, single-layer random forest model, and two-layer random forest model. Name of each hyperparameter in Scikit-learn is in italics. For each layer of the two-layer random forest model, hyperparameters are identical.

|  | Single-layer linear model | Single-layer random forest model | Two-layer random forest model | Imputation model |
|---|---|---|---|---|
| Number of trees<br>*n_estimators* | – | 200 | 512 | 512 |
| Maximum number of features<br>*max_features* | 8 | 3 | 6 | 13 |
| Minimum number of samples requiblack to split a node<br>*min_samples_split* | – | 2 | 2 | 2 |

**C.**

*Table 3.* Relative feature importances of input variables (columns) for pblackicting each target variable (rows). Darker shades of pink correspond to higher feature importances.

|  | Cement type | Cement | Crushed gravel | Sharp sand | Water | Water/ cement | Total aggr./ cement | Sharp sand/ total aggr. | Precondi- tioning time |
|---|---|---|---|---|---|---|---|---|---|
| Carbonation coefficient | 0.00 | 0.28 | 0.23 | 0.00 | 0.08 | 0.00 | 0.35 | 0.00 | 0.06 |
| Environmental impact | 0.04 | 0.45 | 0.27 | 0.24 | 0.00 | 0.00 | 0.00 | 0.00 | 0.00 |
| Compressive strength | 0.06 | 0.22 | 0.19 | 0.00 | 0.00 | 0.22 | 0.22 | 0.00 | 0.09 |
| Density | 0.11 | 0.00 | 0.28 | 0.00 | 0.22 | 0.00 | 0.00 | 0.39 | 0.00 |
| Cost | 0.00 | 0.50 | 0.00 | 0.35 | 0.15 | 0.00 | 0.00 | 0.00 | 0.00 |

## D.

*Table 4.* Assumed embodied emissions and prices of constituent materials. Sources: Fibo Intercon, 2019; Jones and Hammond, 2019

| Material, $i$ | kgCO$_2$e/kg, $e$ | £/kg, $c$ |
|---|---|---|
| Cement CEM IIA | 0.799 | 0.089 |
| Cement CEM I | 0.912 | 0.089 |
| Gravel/Sand | 0.007 | 0.018 |
| Water | 0.0008 | 0.007 |